\documentclass[conference]{IEEEtran}
\IEEEoverridecommandlockouts
\usepackage{cite}
\usepackage{multirow, booktabs}
\usepackage{amsmath,amssymb,amsfonts,mathdots}
\usepackage{algorithmic}
\usepackage{graphicx,subfigure}
\usepackage{textcomp}
\usepackage{xcolor}
\usepackage{verbatim, framed}
\usepackage{tikz}
\usetikzlibrary{arrows,shapes}
\def\BibTeX{{\rm B\kern-.05em{\sc i\kern-.025em b}\kern-.08em
    T\kern-.1667em\lower.7ex\hbox{E}\kern-.125emX}}

\newcommand{\matrnn}{\texttt{MAT-RNN}}
\newcommand{\mwtrnn}{\texttt{MAT-RNN}}
\newcommand{\wtternn}{\texttt{WTTE-RNN}}
\newcommand{\sqrnn}{\texttt{SQ-LOSS}}
\newcommand{\rngf}{\texttt{RNG-F}}
\newcommand{\tse}{\text{tse}}
\newcommand{\tte}{\text{tte}}    
    
\begin{document}

\title{Multivariate Arrival Times 
	with Recurrent Neural Networks 
	for Personalized Demand Forecasting}
% Multivariate Arrival Times with Recurrent Neural Networks for Personalized Demand Forecasting

\author{
	\IEEEauthorblockN{Tianle Chen}
	\IEEEauthorblockA{\textit{Department of Statistical Sciences} \\		
	\textit{University of Toronto}\\
	Toronto, Canada \\
	tianle@utstat.utoronto.ca}
	
	\and
	\IEEEauthorblockN{Brian Keng}
	\IEEEauthorblockA{\textit{Data Science} \\
	\textit{Rubikloud Technologies Inc.}\\
	Toronto, Canada \\
	brian.keng@rubikloud.com}

	\and
	\IEEEauthorblockN{Javier Moreno}
	\IEEEauthorblockA{\textit{Data Science} \\
	\textit{Rubikloud Technologies Inc.}\\
	Toronto, Canada \\
	javier.moreno@rubikloud.com}
}

\maketitle

%%%%%%%%%%%%%%%%%%%%%%%%%%%%%%%%%%%%%%%%%%%%%%%%%%%%%%%%%%%%%%%%%%%%%%%%%%%%%%%%
\begin{abstract}
Access to a large variety of data across a massive population
	has made it possible to predict customer purchase patterns
	and responses to marketing campaigns.
In particular, accurate demand forecasts 
	for popular products with frequent repeat purchases
	are essential since these products are 
	one of the main drivers of profits.
However, buyer purchase patterns are extremely diverse and sparse
	on a per-product level
	due to population heterogeneity 
	as well as dependence in purchase patterns
	across product categories.
Traditional methods in survival analysis have proven effective
	in dealing with censored data
	by assuming parametric distributions on inter-arrival times.
Distributional parameters are then fitted,
	typically in a regression framework.
On the other hand, neural-network based models 
	take a non-parametric approach
	to learn relations from a larger functional class.
However, the lack of distributional assumptions
	make it difficult to model partially observed data.
In this paper, we model directly the inter-arrival times
	as well as the partially observed information at each time step
	in a survival-based approach using Recurrent Neural Networks (RNN)
	to model purchase times jointly over several products.
Instead of predicting a point estimate for inter-arrival times,
	the RNN outputs parameters that define a distributional estimate.
The loss function is
	the negative log-likelihood of these parameters
	given partially observed data.
This approach allows one to leverage
	both fully observed data as well as partial information.
By externalizing the censoring problem 
	through a log-likelihood loss function,
	we show that substantial improvements over
	state-of-the-art machine learning methods can be achieved.
We present experimental results based on two open datasets
	as well as a study on a real dataset
	from a large retailer.
\end{abstract}

\begin{IEEEkeywords}
% top5 start
Survival Analysis, 
Time series analysis,
Neural networks, 
Consumer products,
Multivariate statistics,
% top5 end
Maximum likelihood modeling,
Bayesian network models,
Forecasting,
Marketing
\end{IEEEkeywords}

%%%%%%%%%%%%%%%%%%%%%%%%%%%%%%%%%%%%%%%%%%%%%%%%%%%%%%%%%%%%%%%%%%%%%%%%%%%%%%%%
\section{Introduction}
% retail space and data
% Modern retailers have access to 
% 	massive amounts of consumer behavior data
% 	through customer loyalty programs, purchase histories, 
% 	and responses to direct marketing campaigns.  
% These data sources allow retailers to
% 	customize their marketing communications at the individual level 
% 	through personalized content, promotions, 
% 	and recommendations via channels such as email, mobile and direct mail.
Accurately predicting each customer's behavior
	for each product is useful in direct marketing efforts
	which can lead to significant advantages for a retailer 
	by driving increased sales, margin, and return on investment 
	\cite{dma2009}.
Of special interest are replenishable products 
 	such as regularly consumed food products (e.g. milk) 
 	or regularly replenished personal care products (e.g. soap).
These products drive 
	store traffic, basket size, and customer loyalty,
	which are of strategic importance 
	in a highly competitive retail environment
	\cite{mckinsey2016}.

% % considerations for the problem.
% Accurately forecasting individual demand for single products 
% 	is a complex problem with many facets.
% It necessitates predicting not only 
% 	the next most likely time of purchase 
% 	but also having an accompanying measure of uncertainty is desirable
% 	due to the inherent randomness of an individual's behavior.  
% This coincides with the reality of
% 	sparse observations 
% 	(e.g. few observations for individual customers) 
% 	and partial information 
% 	(e.g. purchases of related products and time since last purchase).
% Sequential dependence needs to be modeled because 
% 	future purchase patterns are heavily influenced by past behavior.  
% Additionally, 
% 	strong correlation between purchase patterns across substitutable products 
% 	indicates that customer behavior must also be jointly predicted.
% % example: deli skus!
% For example, among purchasers of items in a basket of 12 deli products 
% 	that we considered, there are 79,980 unique purchasers.
% Purchase histories for any single product is sparse.
% For any single product in this basket,
% 	the average customer buys only between 0.12 to 0.67 items
% 	over a 1.5 year period
% 	but aggregating over all the products in the basket
% 	indicates that these customers purchase on average 3.58 items 
% 	during the same period.
% This is not surprising since people tend to prefer variety in their meals
% 	even though their choice of whether to purchase a deli product
% 	can be much more predictable.

% explain traditional approaches to the problem.
Traditional approaches to this problem 
	defines customers to be ``alive'' if purchases were made.
These models make assumptions on the distributions of
	purchase counts and lifetimes of individual customers
	as well as population heterogeneity
	\cite{cox1984, schmittlein1987, lariviere2004, fader2005}.
They can be framed in a Cox Proportional Hazards model
	where covariates are regressed as a multiplicative factor
	in a baseline hazard function
	\cite{miller1997}.
% For example, 
% 	in the case of the Pareto/NBD model,
% 	an exponential distribution for customer lifetime
% 	is equivalent to a constant hazard function
% 	\cite{schmittlein1987}.
Solving the maximum likelihood problem
	yields optimal distributional estimates
	that model these behaviors.
% These models have been widely deployed
% 	due to their probabilistic approach in dealing with censored data.
In a scenario with sparse purchase data, 
	this can be useful since non-purchases can reveal information 
	about whether a customer is likely to make a purchase in the future.

% explain traditional approaches to the problem are bad
However, these models impose strict assumptions 
	such as independence and stationarity.
Covariates are often modeled linearly,
	further restricting the space
	of possible functional relations that are possible.
While these assumptions were essential for tractability purposes,
	they can no longer be taken for granted when we wish to model
	highly-correlated, high-dimensional and heterogeneous processes.

% recent rnn approaches
More recently, Recurrent Neural Networks (RNNs)
	were applied in predicting arrival times 
	\cite{hung2006, choi2015, joshi2006}.  
They leverage the capacity of RNNs to model sequential data 
	with complex temporal dependencies as well as non-linear associations 
	\cite{graves2013, schmidhuber2015}.
However, these models do not deal explicitly 
	with the uncertainty of random arrival times 
	and are not able to properly exploit censored data.

% In this paper, we integrate 
% 	a survival analysis approach with Recurrent Neural Networks (RNN)
% 	to model purchase times jointly over several products.
In this work, we integrate a probabilistic approach 
	to model partially observed data with RNNs
	in a novel sequence-to-sequence approach 
	to predict multiple inter-purchase times for each customer.
The RNNs predict a sequence of distribution parameters 
	for a random series of ``times to next purchase''
	(i.e. inter-arrival times).
At each time step, 
	we either observe complete information when an arrival has occurred 
	or incomplete information in a period of non-arrival.
This induces a conditional distribution 
	on the partially observed ``inter-arrival time'',
	which allows us to maximize a likelihood function
	to obtain optimal RNN parameters.
The Multivariate Arrival Times Recurrent Neural Network model
	will be referred to as \matrnn\ in this paper.

% Describe experiments done to show efficacy 3-5 sentences.
The efficacy of this approach is shown through experiments 
	performed on data from several benchmark datasets
	and a large retailer.
We see that our model out-performs 
	other state-of-the-art machine learning approaches 
	in predicting whether a customer made purchases in the next time period.
The results show that \matrnn\ performs better
	in the ROC-AUC metric \cite{powers2011}
	(average ROC-AUC over per-product predictions)
	in 4 out of the 5 categories of products considered.
Additionally, results on the benchmark and synthetic datasets 
	show comparable performance increases 
	when compared to traditional survival model techniques
	and RNNs trained on the usual squared-loss metric.
Implementations and comparisons on open datasets
	will be published on a publicly-accessible repository.

\section{Related Work}
% rnn based approaches
% rnn+survival approaches
% rnn+seq2seq likelihood approaches

% Survival approaches have been well explored 
% 	and have been used extensively
% 	to predict customer purchases and churn
% 	\cite{cox1984, schmittlein1987, lariviere2004, fader2005}
% 	by modeling a customer ``lifetime''
% 	where actively purchasing customers are deemed to be ``alive''
% 	and make purchases independently.
% Heterogeneity of population purchase rates
% 	and population ``lifetimes''
% 	are assumed to follow parametric distributions.
% These models were developed to model a customer's purchases
% 	assuming that it can be sufficiently described by a few distributions.
% This survival framework is well-suited to our problem 
% 	of predicting the time to next purchase.
% However, incorporating covariates in this context
% 	generally requires imposing a linearity assumption 
% 	on one or a few of the model parameters for tractability,
% 	but is an unrealistic assumption if the number of data features is large.

Machine learning approaches to demand forecasting 
	have gained popularity in recent years.
Random Forest models and other ensemble methods in particular 
	have been widely deployed
	and have enjoyed success with binary predictions
	due to their scalability to wide datasets,
	ease of training and regularization strategies
	\cite{berry2004, rafet2015}.
However, tree-based methods are difficult 
	to extend to modeling partial information
	when dealing with sparsely observed inter-arrival times.
An approach dealing specifically with survival is DeepSurv
	\cite{deepsurv},
	where the output of a neural net is fed into a risk function
	and used in a Cox Proportional Hazards (CPH) model.
A CPH model is a survival model where
	the actual hazard rate for each individual 
	is assumed to be a multiplicative factor
	of a common hazard function
	\cite{klein2003}.
Even so, the common roadblock to both of these models
	is that they are unable to use and model
	sequentially-dependent covariates
	and survival statuses.

\subsection{Survival Analysis and RNNs}

Recurrent Neural Nets (RNN) have been proven 
	to model data with complex sequential dependencies
	\cite{graves2013}.
Their success hinged on the development of a 
	Long Short Term Memory (LSTM) structure
	\cite{hochreiter1997}
	that incorporates gates to recursively update an internal state
	in order to make sequential path-dependent predictions.
A common method in recent years
	is to train RNNs to make point estimates for time-to-event 
	by minimizing a distance-based metric
	\cite{hung2006, choi2015, joshi2006}.
The downside is that unobserved arrival times
	cannot be explicitly accounted for in these models.
However, non-arrivals can reveal a significant amount of information
	and is not currently exploited by these RNN implementations.

Combining survival analysis with RNNs has been explored recently
	in order to exploit time series covariates.
One approach uses RNNs as a feature extraction step
	that feeds into a survival model
	\cite{liao2016}.
The RNN takes covariates and 
	the sequential outputs are pooled 
	and incorporated as factors in a Cox Proportional Hazard model.
The model is then fitted end-to-end
	so that RNN parameters and hazard parameters
	are jointly trained.
Experiments showed that this approach achieves an acceptable result
	in failure time prediction.
However, modeling survival status 
	as developing with covariates concurrently
	is a more general approach and possibly more realistic as well.

\subsection{Sequence-to-Sequence Approach to Model Arrival Times}

% covariates can be modelled concurrently with survival status
A recent paper described a framework 
	for using likelihood-based losses in RNNs 
	to model sequential data in a sequence-to-sequence manner
	\cite{flunkert2017}.
A naive approach using this method 
	requires that we model arrivals as a point process,
	where arrival intensity models only the arrival counts per unit time.
This cannot be adapted easily to predict purchases
	by individual customers.
A variation of this is \wtternn\ (Weibull Time-To-Event RNN),
	which models the inter-arrival times directly
	by minimizing a likelihood function
	that utilizes partial information
	\cite{martinsson2016}
	and achieves satisfactory results
	in predicting univariate arrival processes.
However, the theoretical framework proposed
	makes it hard to extend to modeling multivariate arrival processes.
It also assumes a memoryless arrival process,
	which the author has recognized is unrealistic.
Even so, this approach highlights the fact that 
	there can be advantages to viewing 
	this arrival times prediction problem
	in the context of survival analysis
	and in using RNNs to predict distributional parameters
	in a sequence-to-sequence approach.
The proposed theoretical framework (\matrnn) 
	addresses these problems
	and shows that it can be extended easily 
	to model Multivariate Arrival Times more generally.
Not only can this be used in a survival context
	where there is only a single arrival time 
	(i.e. death or failure),
	but can also be used in a multiple arrivals setting 
	(i.e. purchases).

\section{A Multivariate Arrival Times Recurrent Neural Net Model}
% This first paragraph should just give an overview of the method in plain English, then let's get into the math in the sub sections.  Like:
% - To model inter-purchase time of a single customer, we use an RNN, where at each time step, it outputs a set of parameters for a distribution of the time to next arrival
% - At each time step, we have a binary indicator observation z_t which indicates a purchase or no purchase
% - The fitting process aims to minimize the likelihood of each distribution (one for each time step) by fitting the distribution to the next time to arrival.
% - Probably should mention that we use a Weibul
% Something like that, relatively simple language that gives the reader a mental model of how it all works.  
% The Figure should be the analogue to this description.  
% So basically in 1/2 page, they can quickly get an idea of what your method is without reading all the math.

The proposed model specifies a Recurrent Neural Net (RNN)
	to output distributional parameters
	which represent predictions 
	for the remaining time to arrival.
By iterating through time for each customer,
	the RNN outputs sequential distributional estimates
	for the remaining time until the next purchase arrival,
	giving a personalized demand forecast.
In this model, for each customer and product pair,
	every inter-arrival time
	is assumed to be a realization 
	of a distinct random variable.
It can be dependent on other product purchases
	as well as on earlier purchases of the same product.
These inter-arrival times
	are assumed to be independent 
	conditioned on observing a latent state.

At each time step,
	the random variable observed is not the inter-arrival time,
	but the remaining time to next arrival,
	since we have observed partial information
	(i.e. the lower bound of the next inter-arrival time).
This remaining time to next arrival
	is often referred to as a conditional excess random variable
	[Section \ref{condexcessrv}],
	whose distribution is used to derive the log-likelihood
	at each time step
	[Section \ref{llikecomp}].
We assume that our observed process
	follows a conditional independence structure
	where these conditional excess random variables 
	are assumed to be independent
	given the internal state of the RNN
	[Section \ref{distrnn}, \ref{condindep}].
The loss function is defined to be the negative log-likelihood
	and the optimal RNN parameters under such a setup
	generate distributional parameters which are most likely 
	to explain the observed data.
Hence, RNN outputs at the end of training period
	are our best distributional estimates 
	for the remaining time to next purchase.

\subsection{\label{condexcessrv} The Conditional Excess / Remaining Lifetime Random Variable}

We will denote the random variable representing
	the remaining time till next arrival
	conditioned on the current information
	as $Z_t$.
This random variable is not the true inter-arrival time,
	but is instead a version 
	that is conditioned on observing partial information.
In a survival analysis framework
	where there is only one inter-arrival time (i.e. the time of death),
	this is the remaining lifetime.

Consider an arrival process 
	where $W_n$ is the time of the $n$-th arrival
	and let $W_0 = 0$, the start of training period.
Let $N(t)$ be the number of arrivals by time $t$
	[Equation \ref{eqn:numarrivalsbyt}].
Also let $Y_n$ be the inter-arrival time of the $n$-th arrival,
	which is the difference between consecutive arrival times
	[Equation \ref{eqn:iatn}].
\begin{equation}
N(t) = \max \{n\ |\ W_n \leq t\}
\label{eqn:numarrivalsbyt}
\end{equation}
\begin{equation}
Y_n = W_n - W_{n-1}
\label{eqn:iatn}
\end{equation}

In the context of our problem,
	at a particular time $t$,
	the number of arrivals observed is $N(t)$
	and we wish to predict 
	the subsequent (i.e. the $\{N(t)+1\}$-th arrival)
	and its inter-arrival time $Y_{N(t)+1}$.
Let $\tse(t)$ (time-since-event)
	be the amount of time that has elapsed 
	since the last arrival
	or start of training period,
	whichever is smaller
	[Equation \ref{eqn:tse}].
This represents the censoring information
	that is available to the RNN
	at each time $t$.
Define $\tte(t)$ (time-to-event) 
	be the amount of time remaining
	until the next arrival or the end of testing period ($\tau$),
	whichever is smaller
	[Equation \ref{eqn:tte}].
\begin{equation}
\tse(t) = t - W_{N(t)}
\label{eqn:tse}
\end{equation}
\begin{equation}
\tte(t) = \min \{ W_{N(t)+1} - t, \tau - t \} 
\label{eqn:tte}
\end{equation}

\begin{figure}[htbp]
	\centering
	\subfigure{\includegraphics[height=4cm]{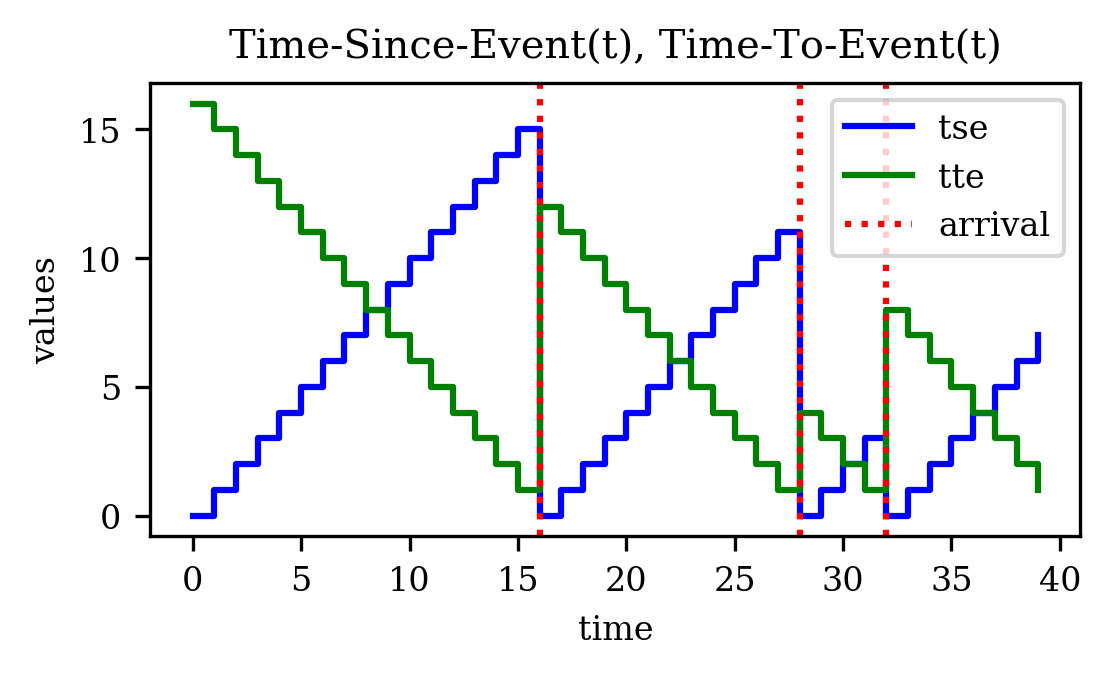}}
	\caption{Notation for $\tse(t), \tte(t)$ with three arrivals:
		The arrival times are at 16, 28 and 32.}
	\label{fig:notationaids}
	\end{figure}

Consider a sample process with three arrivals where 
	$W_1 = 16, W_2 = 28, W_3 = 32$,
	such that 
	$Y_1 = 16, Y_2 = 12, Y_3 = 4$.
Also, $N(t)$ is a piecewise constant function
	which is $0$ for before the first arrival at $t=16$
	and jumps by 1 at each arrival time $W_n$.
We plot $\tse(t), \tte(t)$
	for $t$ until $\tau = 40$, which is the end of the training period
	[Figure \ref{fig:notationaids}].

The remaining time until next arrival ($Z_t$)
	is a conditional random variable
	that depends only on $Y_{N(t)+1}$,
	which is the inter-arrival time of the subsequent arrival.
We can thereby define this random variable 
	given the observed information
	[Equation \ref{eqn:conditionalexcess}],
	which is commonly known as a conditional excess random variable
	\cite{finkelstein2008}.
\begin{equation}
Z_t = Y_{N(t)+1} - \tse(t)\ |\ Y_{N(t)+1} > \tse(t).
\label{eqn:conditionalexcess}
\end{equation}

In our approach, we do not model the distribution of $Z_t$ directly.
Instead, the RNN predict parameters for $Y_{N(t)+1}$
	and the losses are computed based on the distribution of $Z_t$,
	which has a distribution induced by $Y_{N(t)+1}$
	based on partial information (i.e. $\tse(t)$)
	\footnote{In \wtternn,
		$Z_t$ is formulated as $Y_{N(t)+1} - \tse(t)$
		which are conditionally independent in time.
	The parameters which are outputs of the RNN in \wtternn\
		describe the distribution of $Z_t$
		\cite{martinsson2016},
		not $Y_{N(t)+1}$ as is the case in our approach.
	We find that modeling the true lifetime
		and computing losses based on the remaining lifetime
		is a preferable approach
		which lends direct analogy to survival methods
		and interpretability.}.
This additional structure helps to simplify the incomplete information problem.

For example, consider $Z = Y-t\ |\ Y > t$.
This random variable $Z$ 
	is conditioned on the fact that $Y$ has been observed to exceed $t$
	and we are interested in the excess value
	(i.e. $Y-t$).
It is clear that the distribution of $Y$ induces a distribution on $Z$
	[Equation \ref{eqn:condiexcesssurv}].
\begin{equation}
P(Z > s) = P( Y-t > s | Y > t) 
	= \frac{P(Y > s+t)}{P(Y > t)}.
\label{eqn:condiexcesssurv}
\end{equation}

\subsection{\label{llikecomp} Log-Likelihood Computation}

There are two cases to properly define the log-likelihood function.
When the next arrival time is observed,
	the likelihood evaluation is $P(Z_t \in [\tte(t), \tte(t)+1])$,
	since inter-arrival times are only discretely observed.
However, where the time to next arrival is not observed
	(i.e. no more subsequent arrivals are observed by end of training),
	the likelihood evaluation is instead $P(Z_t > \tte(t))$,
	namely the survival function.
Therefore at each time $t$,
	the random variable $Y_{N(t)+1}$ 
	which has distribution parametrized by $\theta_t$,
	induces a distribution on $Z_t$.
Thus the log-likelihood at each time $t$
	can be written as follows
	[Equation \ref{eqn:loglikelihood}].
\begin{equation}
l_t(\theta_t) = 
	\begin{cases}
    	\log P(Z_t \in [\tte(t), \tte(t)+1]) & \text{if uncensored}\\
    	\log P(Z_t > \tte(t)) & \text{otherwise}
    	\end{cases}
\label{eqn:loglikelihood}
\end{equation}

In survival analysis terminology,
	we recall that $Z_t$ is the remaining lifetime random variable
	(i.e. time to death conditioned on current age).
For the uncensored case, the likelihood is the death probability
	and in the right-censored case, it is the survival probability.

\begin{figure}[h]
	\centering
	\subfigure
		[If next arrival is observed then log-likelihood is density]
		{\includegraphics[height=4cm]{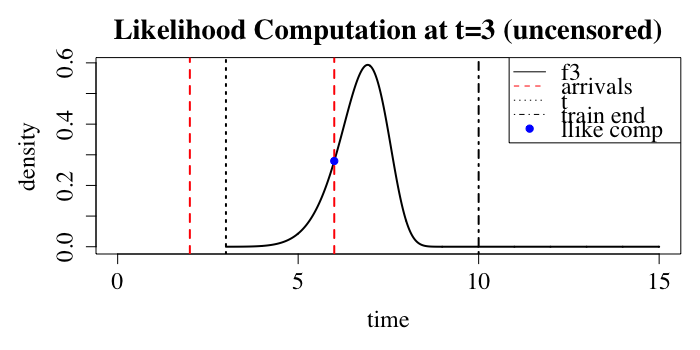}}
	\subfigure
		[If next arrival is unobserved then log-likelihood is survival.]
		{\includegraphics[height=4cm]{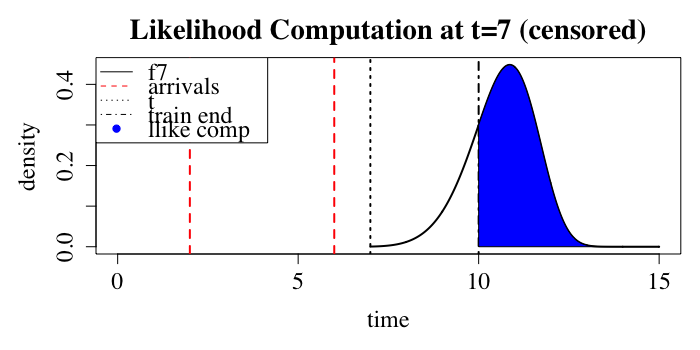}}
	\caption{Log-Likelihood Visualization for Different Censoring Status}
	\label{fig:loglike_viz}
	\end{figure}

We plot distributional estimates
	at two times to illustrate these cases
	[Figure \ref{fig:loglike_viz}].
In the uncensored log-likelihood computation,
	we assume that \texttt{f3}
	is the density function of $Z_3$,
	which is the predictive distribution 
	for the remaining time until next arrival
	at time step 3.
Since the next arrival is observed to have occurred at time 6,
	the remaining time to next arrival is 3,
	so we evaluate \texttt{f3} at the value 3.
In the censored case,
	we consider the predictive distribution 
	for remaining time until next arrival at time step 7.
We note that the next arrival was not observed
	by end of training period at time step 10
	hence the right tail of $Z_7$
	(i.e. $\geq 3$)
	was used to compute the log-likelihood.

It is essential that the class of distributions for $Y_n$ 
	is assumed to have differentiable and numerically stable forms 
	for the density and survival functions
	to exploit the back-propagation algorithm
	for efficient gradient computation
	\cite{hochreiter1997}.
An example is the Weibull distribution
	parametrized by scale ($\lambda$) and shape ($k$),
	which is used in our approach,
	whose survival function is 
	made up of only \texttt{exp()}
	and \texttt{power()} functions
	[Equation \ref{eqn:weibullsurv}].
The details of the conditional excess random variables
 	based on Weibull distributions can be found in the Appendix
 	[Section \ref{appendix:weibull_survival}].
\begin{equation}
S(y) = P(Y>y) = e^{-(y/\lambda)^k}
\label{eqn:weibullsurv}
\end{equation}

\subsection{\label{distrnn} An RNN with distributional outputs}

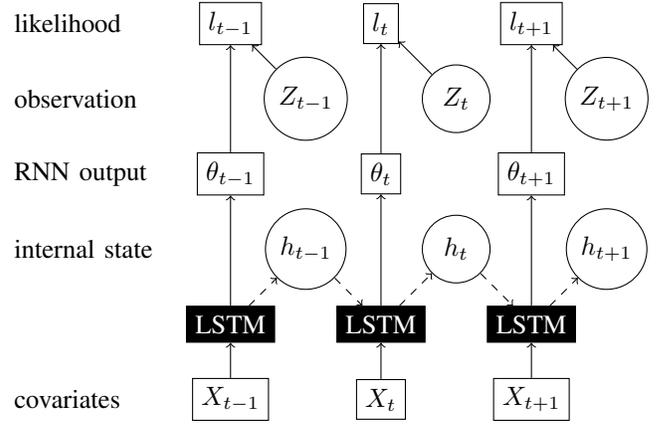
\begin{figure}[h]
\centering
\begin{tikzpicture}
	\node (covariates)     [anchor=west] at (0, 0) {covariates};
	\node (internal state) [anchor=west] at (0, 2) {internal state};
	\node (rnn output)     [anchor=west] at (0, 3) {RNN output};
	\node (observation)    [anchor=west] at (0, 4) {observation};
	\node (likelihood)     [anchor=west] at (0, 5) {likelihood};

	\tikzstyle{every node} = [draw, shape=rectangle, fill=black]
	\node (lstmprev) at (3, 1) {\textcolor{white}{LSTM}};
	\node (lstmnow)  at (5, 1) {\textcolor{white}{LSTM}};
	\node (lstmtmr)  at (7, 1) {\textcolor{white}{LSTM}};

	\tikzstyle{every node} = [draw, shape=rectangle]
	\node (llikeprev) at (3, 5) {$l_{t-1}$};
	\node (llikenow)  at (5, 5) {$l_t$};
	\node (lliketmr)  at (7, 5) {$l_{t+1}$};
	\node (xprev)     at (3, 0) {$X_{t-1}$};
	\node (xnow)      at (5, 0) {$X_t$};
	\node (xtmr)      at (7, 0) {$X_{t+1}$};
	\node (abprev)  at (3, 3) {$\theta_{t-1}$};
	\node (abnow)   at (5, 3) {$\theta_t$};
	\node (abtmr)   at (7, 3)  {$\theta_{t+1}$};

	\tikzstyle{every node} = [draw, shape=circle, minimum size=0.9cm]
	\node (obsprev) at (4, 4) {$Z_{t-1}$};
	\node (obsnow)  at (6, 4) {$Z_t$};
	\node (obstmr)  at (8, 4) {$Z_{t+1}$};
	\node (zprev)     at (4, 2) {$h_{t-1}$};
	\node (znow)      at (6, 2) {$h_t$};
	\node (ztmr)      at (8, 2) {$h_{t+1}$};

	\foreach \from/\to in 
		{xprev/lstmprev, lstmprev/abprev,
			xnow/lstmnow, lstmnow/abnow,
			xtmr/lstmtmr, lstmtmr/abtmr,
			obsprev/llikeprev, abprev/llikeprev,
			obsnow/llikenow, abnow/llikenow,
			obstmr/lliketmr, abtmr/lliketmr}
		\draw [->] (\from) -- (\to);
	\foreach \from/\to in
		{lstmprev/zprev, lstmnow/znow, lstmtmr/ztmr,
			zprev/lstmnow,
			znow/lstmtmr}
		\draw [dashed, ->] (\from) -- (\to);
	\end{tikzpicture}
\caption{RNN Computational Flow:
	Outputs ($\theta_t$) are generated by an LSTM.
	Log-likelihoods at each time are computed 
		as log of densities parametrized by $\theta_t$,
		evaluated at $z_t$}
\label{fig:rnn_graph}
\end{figure}

A particular RNN structure that had proven to be successful
	in modeling sequential data is Long Short Term Memory (LSTM)
	\cite{hochreiter1997},
	although it can be replaced by other RNN implementations.
At each time ($t$),
	the outputs of the LSTM,
	which is parametrized by $\Theta$,
	are passed through an activation function
	so that they are valid parameters
	of a distribution function ($\theta_t$).
Then, the log-likelihood is computed for each time step ($l_t$)
	as defined earlier
	[Equation \ref{eqn:loglikelihood}].
The computational graph is shown [Figure \ref{fig:rnn_graph}],
	where $h_t$ is the internal state of the LSTM
	and $X_t$ are the covariates at each time $t$.

The loss under such a set-up is the negative of the log-likelihood.
Optimal parameters for the LSTM ($\Theta$) are ones that
	output a series of distributional estimates $\theta_t$
	that best ``explain'' the sequence of data observed.
In the event of an uncensored arrival time at time $t$,
	the optimal choice of weights $\theta_t$ is one
	that generates a density that has a peak 
	close to the actual arrival time.

\subsection{\label{condindep} Computing Overall Loss through Conditional Independence}

\begin{figure}[h]
\centering
\begin{tikzpicture}
	\node (hidden state) [anchor=west] at (0, 0)   {hidden state};
	\node (observation)  [anchor=west] at (0, 1.5) {observation};

	\tikzstyle{every node} = [draw, shape=circle, minimum size=1cm]
	\node (aprev) at (3, 0)   {$h_{t-1}$};
	\node (wprev) at (3, 1.5) {$Z_{t-1}$};

	\node (anow)  at (5, 0)   {$h_t$};
	\node (wnow)  at (5, 1.5) {$Z_t$};

	\node (atmr)  at (7, 0)   {$h_{t+1}$};
	\node (wtmr)  at (7, 1.5) {$Z_{t+1}$};
	\foreach \from/\to in 
		{aprev/wprev, anow/wnow, atmr/wtmr}
		\draw [->] (\from) -- (\to);
	\foreach \from/\to in
		{aprev/anow, anow/atmr}
		\draw [->] (\from) -- (\to);
	\end{tikzpicture}
\caption{Bayesian Network: Observations 
	are independent conditioned on hidden states}
\label{fig:bayesian_network}
\end{figure}
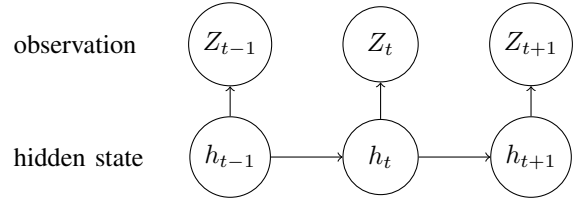

We assume a Bayesian Network 
	similar to a Hidden-Markov model,
	where random variables at each time $t$
	are emitted from a hidden state $h_t$
	[Figure \ref{fig:bayesian_network}].
Recall that $h_t$ represents 
	the internal state of the RNN at each time $t$
	[Figure \ref{fig:rnn_graph}],
	and $Z_t$ is the observed time series.

We can therefore factor the joint distribution of $\{Z_t\}$,
	giving the log-likelihood computation
	for the entire time series as a sum of log-likelihoods at each time,
	such that we obtain the sum below,
	for arbitrary events $E_t$.
Since $E_t$ is determined by the censoring status
	[Equation \ref{eqn:loglikelihood}],
	where $E_t = \{[\tte(t), \tte(t)+1]\}$ if uncensored
	and $E_t = \{> \tte(t)\}$ otherwise,
	we can decompose the overall log-likelihood as a sum
	[Equation \ref{eqn:llikesum}].
\begin{equation}
\begin{aligned}
P(\{Z_t \in E_t\}_{t=1}^\tau |\ \{h_t\}_{t=1}^\tau) 
	& = \prod_{t=1}^\tau P(Z_t \in E_t |\ h_t) \\
l(\{\theta_t\}) 
	& = \sum_t l_t(\theta_t) 
\label{eqn:llikesum}
\end{aligned}
\end{equation}

Assuming that the LSTM/RNN 
	is parametrized by $\Theta$,
	we note that there exists a function $g$
	that recursively maps $X_t$ to $(\theta_t, h_t)$
	that depends only on $\Theta$
	[Equation \ref{eqn:recurrentmapg}].
By substituting $h_{t-1}$, we can write 
	$l_t(\theta_t) = l_t(g_t(\Theta))$
	where $g_t$ depends only on $\Theta, g, \{X_l\}_{l\leq t}$.
Then since the overall log-likelihood is a sum of $l_t(\theta_t)$,
	it can be written as a function of only 
	the RNN parameters ($\Theta$) and observed data.
The structure of the RNN and the back-propagation algorithm
	allows us to compute gradients of any order efficiently
	and therefore find the $\hat{\Theta}$, 
	the minimizer of the overall observed loss
	\cite{hochreiter1997}.
\begin{equation}
(\theta_t, h_t) = g(h_{t-1}, X_t\ |\ \Theta)
\label{eqn:recurrentmapg}
\end{equation}

\subsection{\label{activation} Activation Functions for Distributional Parameters}

The neural net outputs must be transformed 
	such that they are parameters of a distribution.
In our case, we used a Weibull distribution,
	which is parametrized by shape and scale parameters,
	both of which are positive values.
We initialized the RNN output for scale at the 
	maximum-likelihood estimate (MLE) 
	for the scale parameter of a Weibull distribution
	whose shape parameter is 1
	as this was found to be useful 
	in preventing likelihood-evaluation errors
	\cite{martinsson2016}.
We also chose a maximum shape parameter (set at 10)
	and pass the RNN output for shape
	through a sigmoid function,
	which is rescaled and shifted such that 
	$\sigma^*:\mathbb{R}\rightarrow (0, 10)$ 
	and 
	$\sigma^*(0) = 1$.
For the scale parameter, an exponential function is used,
	which is rescaled such that it maps 0 to 
	the average inter-arrival-time.

\subsection{Extension to Multi-Variate Waiting Times}

To model multivariate arrivals, 
	we assume there are $p$ different 
	arrival processes of interest.
For the $i$-th waiting time of interest,
	we define $W_{i, n}$ to be the time of the $n$-th arrival 
		of this type
	and $N_i(t), Y_{i, n}$ be likewise defined.
We also need to define 
	$\tse(i, t), \tte(i, t)$
	to be that for the $i$-th type as well.
Similarly define its associated 
	conditional excess random variable
	[Equation \ref{eqn:conditionalexcess_multivar}].
\begin{equation}
Z_{i, t} = Y_{i, N_i(t)+1}-\tse(i, t)\ |\ Y_{i, N_i(t)+1}>\tse(i, t)
\label{eqn:conditionalexcess_multivar}
\end{equation}

In the earlier framework 
	[Figure \ref{fig:bayesian_network}], 
	we can let 
	$Z_t = [Z_{1, t}, \ldots, Z_{p, t}]$
	and let the RNN output 
	$\theta_t = [\theta_{1, t}, \ldots, \theta_{p, t}]$.

Then we write the log-likelihoods for each event type where
	$l_{i, t}(\theta_{i, t}) = \log P(Z_{i, t} = \tte(i, t))$
	or $l_{i, t}(\theta_{i, t}) = \log P(Z_{i, t} > \tte(i, t))$,
	recalling that the former is for the case where 
		the next arrival is observed
	while the latter is for the case where
		the no arrivals are observed until the end of training.

The earlier Bayesian Network structure
	[Figure \ref{fig:bayesian_network}]
	requires minimal modifications
	as we merely require that the emissions are conditionally independent
	given $h_t$.
This then allows us to compute the log-likelihood at each time as a sum,
	$l_t(\theta_t) = \sum_i l_{i, t}(\theta_{i, t})$.
Since the LSTM network is still parameterized by $\Theta$,
	the remaining operations are exactly the same as earlier.
In this way, temporal dependence 
	as well as dependence between the $p$ arrival processes
	can be modeled by the RNN,
	whose weights $\Theta$ will then be optimized by training data.
This allows us to model other outputs as well 
	by appending
	$[K_{1, t}, \ldots, K_{p, t}]$ to $Z_t$
	where $K_{j, t}$
	is some other variable of interest
	for process $j$ at time $t$.

\subsection{Masking of Non-Inter-Arrival-Times}

In multi-variate purchase arrival times,
	we found that masking sequences observed 
	before the first arrival of each product
	is useful in preventing numerical errors
	encountered in stochastic gradient descent.
As such, log-likelihoods computed for
	time steps before the earliest arrival
	are masked.
This ensures that RNN parameters
	are not updated due to losses incurred
	during these times.

\subsection{\label{predictions} Predicting Time to Next Arrival}
	
Predictions are simple to compute
	since at each time $t$, 
	we can use the estimated parameter $\theta_t$
	to compute the expectation of any function of $Z_t$,
	assuming that $Z_t$ is distributed according to $\theta_t$.
Since we are concerned with the next arrival time
	after the end of training period 
	(time $\tau$),
	we can compute many different values of interest.

For example,
	if we want to find the predicted probability 
	that the next arrival 
	will occur within $\gamma$ time after end of training, 
	we can compute $P(Z_\tau \leq \gamma)$.
We can also compute a deferred arrival probability,
	which is the probability that 
	the next arrival will occur within an interval between
		$\gamma_1$ and $\gamma_1 + \gamma_2$ time 
		after end of training
		given that we know it will not occur
		within $\gamma_1$ time after the end of training.
This can be found by computing 
	$P(Z_\tau \in [\gamma_1, \gamma_1+\gamma_2]\ |\ Z_\tau > \gamma_1)$.
The quantities of interest may not necessarily be limited to probabilities
	(e.g. mean, quantiles of the predictive distribution)
	and can be extended to generate other analytics
	for revenue analysis or forecasting
	that depends on the subsequent purchase time.

% {\color{red} Consider re-run quantiles for a visualization example...}

% \begin{figure}[h!]
% 	\centering
% 	\includegraphics[height=4cm]{pics/eg_quantiles.png}
% 	\caption{Example of In-Sample Predictions for 
% 		Remaining Time to Arrival}
% 	\label{fig:insample-eg}
% \end{figure}

% We can visualize in-sample predictions of our model.
% Quantiles are plotted 
% 	for the remaining time to arrival $Z_{i, t}$
% 	using the parameters output from the RNN at each time step
% 	to make in-sample predictions for a purchase arrival processes
% 	for an individual
% 	[Figure \ref{fig:insample-eg}].
% We note that the in-sample predicted time to next event
% 	does not necessarily follow the actual time to next event
% 	since significant variation is to be expected among individuals.
% Best estimates for time to purchase
% 	is represented by the distribution of $Z_{i, \tau}$,
% 	which is summarized by the quantiles 
% 	at the last time step (i.e. weeks = 78).

\section{Experimental Results}
We ran experiments to show the efficacy of the proposed model.
Since our model provides distributional estimates for time to next event,
	we can evaluate its performance
	in a binary classification framework
	as well as in a regression context.
This is explored by running experiments on two open datasets.
The first is store purchases so there are multiple arrivals
	but with no covariates.
The second is failure time prediction with many covariates.
Finally, we apply the model to a real dataset of retail purchases
	in order to compare real-world performances
	of the considered models.
We were unable to find open data for multivariate arrival times
	with a large number of arrival types
	but the publicly available code 
	can be easily extended to run on such data.

\subsection{Performance Evaluation}

Binary classification performance is evaluated using the
	Receiver Operating Characteristics Area Under Curve (ROC-AUC),
	which is equivalent to the Concordance Index (C-Index)
	commonly used in survival analysis
	\cite{harrell1984}.
Point prediction performance is evaluated using
	specified distance metrics.

\subsection{\label{nnetspecs} Recurrent Neural Net Structure}

For all experiments in this section,
	a simple structure for our neural net is chosen
	where there are only three stacked layers,
	with two \texttt{LSTM} layers of size $W$
	followed by a densely connected layer of size $2p$,
	where $p$ is the number of arrival processes.
The densely connected layer
	transforms the \texttt{LSTM} outputs
	to a vector of length $2p$.
We chose a Weibull distribution
	as the family of distributions
	for our inter-arrival times,
	which has two parameters,
	namely the scale and shape.
In \matrnn, the outputs are then passed through an activation layer
	[Section \ref{activation}].
For squared-loss RNNs,
	the activation is passed through a softplus layer
	since time to arrivals are non-negative.
A masking layer is applied prior to the other layers
	on time indices prior to the initialization of the time series
	so that losses incurred during those time steps
	do not affect optimization.

\begin{figure}[h]
\centering
\begin{tikzpicture}
	\tikzstyle{every node} = [draw, shape=rectangle]
	%\node (masking)  at (0, 0) {\texttt{Masking}};
	\node (lstm1)     at (0, 0) {\texttt{LSTM}(W)};
	\node (lstm2)     at (2, 0) {\texttt{LSTM}(W)};
	\node (dense)    at (4, 0) {\texttt{Dense}(2p)};
	%\node (activate) at (6, 0) {\texttt{Activation}};
	\foreach \from/\to in 
		{lstm1/lstm2, lstm2/dense} %, dense/activate}
		\draw [->] (\from) -- (\to);
	\end{tikzpicture}
\caption{Recurrent Network Setup for a $p$-variate Arrival Process
	with $W$-long Hidden State}
\label{fig:nnetspecs}
\end{figure}
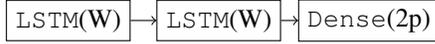

This structure is the same
	for other neural network based models
	used for benchmark comparison.
The networks are trained with \texttt{Adam}
	\cite{adam}
	and non-default training parameters
	are specified where possible.
Gradients are component-wise clipped at $5$
	to prevent numerical issues.
Implementation was done using \texttt{Keras}
	\cite{keras}
	as our \texttt{TensorFlow} 
	\cite{tensorflow}
	wrapper.
Evaluation metrics and ensemble predictors
	are implemented using \texttt{scikit-learn} 
	\cite{scikitlearn}.	
Implementations and comparisons on open datasets
	will be published on a publicly-accessible repository
	\footnote{http://github.com/rubikloud/matrnn}.

\subsection{Comparisons on Open Datasets}

We show the flexibility of the proposed \matrnn\ model 
	with two open dataset benchmarks.
These two problems are often tackled with different models
	since the prediction problem is different.
The proposed \matrnn\ model however,
	can be adapted to solve these problems 
	since they can be modeled by 
	a distributional approach to inter-arrival times.

The first example is the \texttt{CDNOW} dataset
	[Section \ref{cdnow}].
We consider a binary classification problem
	where we predict if purchases are made during a testing period.
Predictions from \matrnn\ can be computed as 
	the probability that the inter-arrival time 
	occurs before the end of the testing period.
Data available is limited to only the transaction history
	(i.e. purchase date, purchase quantity, customer id, etc)
	without other covariate data.
This type of data is particularly suited 
	to the simple Pareto/NBD type models discussed earlier.
We show that \matrnn\ out-performs on a dataset even with no covariates.

The second example is based on the \texttt{CMAPSS} dataset
	[Section \ref{cmapss}]
	where we predict the remaining useful lifetime,
	or the time to failure.
Predictions from \matrnn\ can be computed as the mode, mean 
	or some other function of the inter-arrival time distribution.
The training data is an uncensored time series
	where sensor readings and operational settings are collected
	until the engine fails.
A customized loss function was used to evaluate models
	in the PHM08 competition
	\cite{phm08-results}, 
	which we will include for our evaluation.
Since the training data is fully observed,
	we would expect the RNN model to perform well.

\subsubsection{\label{cdnow} Binary Classification Comparison on CDNOW}

Purchase transactions are available 
	from the \texttt{CDNOW} dataset
	\footnote{http://www.brucehardie.com/datasets/CDNOW\_master.zip}
	\cite{cdnow-dataset}
	where number of customer purchases are recorded.
Only transaction dates, purchase counts
	and transaction value are available as covariates.
Our model is trained on weekly purchases 
	and hidden layers are 1-long
	(i.e. $W=1$ [Section \ref{nnetspecs}]).
Binary classification performance 
	is compared to the Pareto/NBD model,
	which is a classical demand forecasting model
	\cite{schmittlein1987}
	using the \texttt{lifetimes} package 
	\cite{lifetimes-package}.
This dataset is often used as an example
	where Pareto/NBD type models do well
	since there's limited covariate data available
	and there's only a single event type.

\begin{figure}[h]
\centering
\centerline{\includegraphics[height=4cm]{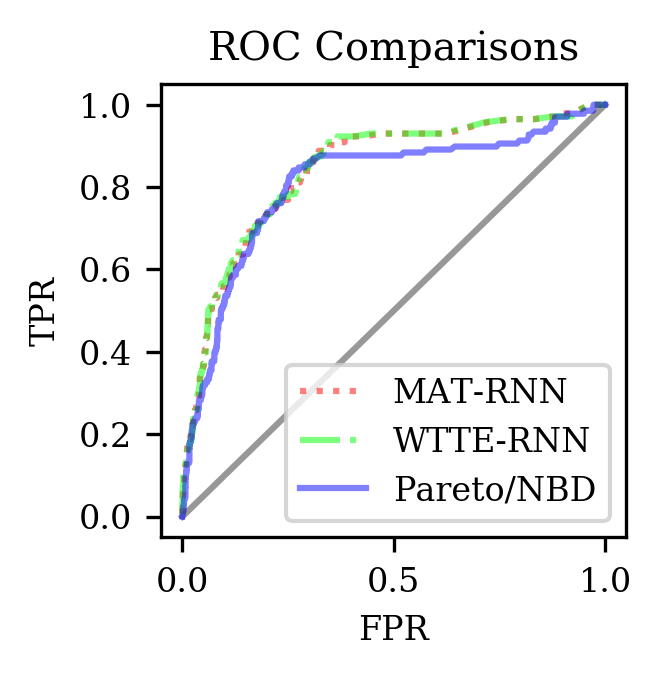}}
\caption{ROC Curves for \texttt{CDNOW} predictions}
\label{fig:testing_cdnow_rocauc}
\end{figure}

With $W=1$, there are only $32$ trainable parameters 
	in the \mwtrnn\ model.
The training period is set at 1.5 years,
	from 1997-01-01
	to 1998-05-31.
Predictions are made for customer purchases
	within a month of the end of training
	(i.e. before 1998-06-30).
We see that \mwtrnn\ achieves an ROC-AUC of 0.84
	which is better compared to 0.80
	that is obtained using the Pareto/NBD estimate
	for the ``alive'' probability
	[Figure \ref{fig:testing_cdnow_rocauc}].
Results from \wtternn\ are similar and 
	yields almost the same ROC-AUC curve.
This is likely due to the mostly memoryless nature 
	of these customer purchases.
It can be seen that this approach of 
	integrating a survival-based
	maximum log-likelihood method with a RNN
	can yield improved prediction accuracy
	even with a small number of weights and on a small dataset.

\subsubsection{\label{cmapss} Point Estimate Comparison on CMAPSS}

The CMAPSS dataset
	\footnote{http://ti.arc.nasa.gov/tech/dash/groups/pcoe/prognostic-data-repository/}
	\cite{cmapss-dataset},
	is a high dimensional dataset on engine performance
	with 26 sensor measurements and operational settings.
In the training model, the engines are run until failure.
In the testing model, data is recorded until a time prior to failure.
The goal is to predict the remaining useful life (RUL)
	for these engines.
We used the first set of engine simulations in the dataset
	(\texttt{train\_FD001.txt})
	which has 100 uncensored time series
	of engines that were run until failure.
The maximum cycles run before failure was found to be at 363.
Time series for each engine was segmented into sliding windows
	of window length 78,
	resulting in 323 windowed time series
	each of length 78.
For the testing dataset, RNN is run on time series 
	78 cycles before end of observation.

A custom loss function was defined 
	for the PHM08 conference competition
	that was based on this dataset,
	where over-estimation is more heavily penalized
	\cite{phm08-results}.
The loss function is defined as follows,
	where $d$ is the predicted RUL subtracted by the true RUL
	[Equation \ref{eqn:phm08loss}].
\begin{equation}
\text{loss}(d) = 
	\begin{cases}
		e^{-d/13} -1
			& d < 0 \\
		e^{d/10} -1
			& d > 0 \\
	\end{cases}.
\label{eqn:phm08loss}
\end{equation}

We compare the performances of the RNN models
	since they are able to make point predictions fro RUL.
The \sqrnn\ model was used with a softplus activation layer
	scaled by average failure time
	and the weights are fitted on squared loss.
The RNN models were trained with $W=64$ 
	[Section \ref{nnetspecs}],
	where there are 2 hidden layers
	each of $W$-long.
A variety of training iterations and learning rates were used,
	and a grid search was performed over all combinations.
Performance is evaluated on the test set
	based on the root mean squared loss metric (rMSE)
	as well as the mean custom loss metric (MCL)
	[Equation \ref{eqn:phm08loss}].
Testing set losses are presented in a table
	[Table \ref{table:cmapssloss}].
We find that \matrnn\ is generally easier to train,
	achieving the smallest losses under almost all hyper-parameters
	other than a slow learning rate and small training iteration.

\begin{table}[h]
\begin{center}
\caption{Comparison on RNN-based Methods for CMAPSS}
\begin{tabular}{|l|l|r|r|r|r|}
\hline
lr   & iters & loss   & \matrnn        & \wtternn        & \sqrnn   \\
\hline
1e-3 & 1e2   & MCL    & \textbf{41.79} & 275.73          & 262.39   \\
     &       & rMSE   & \textbf{32.82} & 41.05           & 42.50    \\
     \cline{2-6}
     & 1e4   & MCL    & \textbf{41.79} & 275.73          & 262.39   \\
     &       & rMSE   & \textbf{32.82} & 41.05           & 42.50    \\
\hline
1e-4 & 1e2   & MCL    & \textbf{45.84} & 355.48          & 446.88   \\
     &       & rMSE   & \textbf{33.16} & 42.10           & 47.53    \\
     \cline{2-6}
     & 1e4   & MCL    & \textbf{41.79} & 275.73          & 262.39   \\
     &       & rMSE   & \textbf{32.82} & 41.05           & 42.50    \\
\hline
1e-5 & 1e2   & MCL    & 1926.13        & \textbf{386.16} & 10041.36 \\
     &       & rMSE   & 53.16          & \textbf{42.04}  & 60.22    \\
     \cline{2-6}
     & 1e4   & MCL    & \textbf{29.34} & 36.84           & 262.39   \\
     &       & rMSE   & \textbf{28.85} & 31.49           & 42.50    \\
\hline
\end{tabular}
\end{center}
\label{table:cmapssloss}
\end{table}

\begin{figure}[h]
	\centering
	\subfigure{\includegraphics[height=4cm]{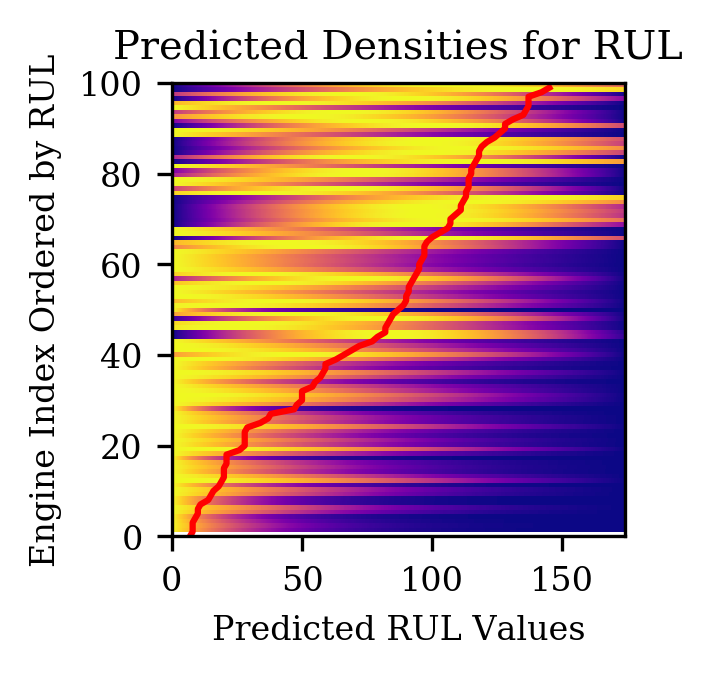}}
	\subfigure{\includegraphics[height=4cm]{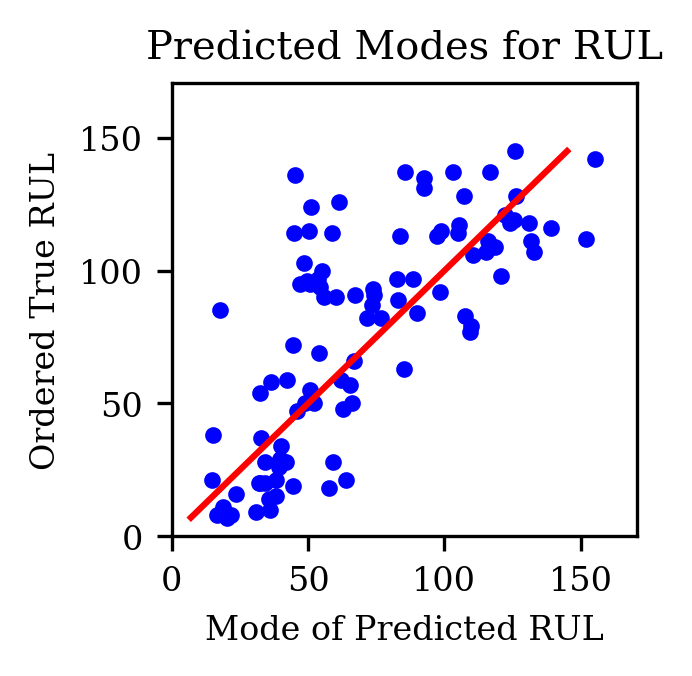}}
	\caption{Predicted RUL Density and Mode on C-MAPSS and True RUL (in red)
		for \matrnn\ with $1e4$ iterations and $1e-5$ learning rate.}
	\label{fig:testing_cmapss_mwtrnnviz}
	\end{figure}

We visualize the predictions of the best performing model,
	achieved by \matrnn\ with $\text{iters}=1e4, \text{lr}=1e-5$
	[Figure \ref{fig:testing_cmapss_mwtrnnviz}].
As expected, we find that the predicted density of the remaining lifetime
	to be highest at the true RUL.
Taking point estimate as the model of the predicted density,
	we can then plot predicted RUL against true RUL,
	which indicates that we have a good predictive accuracy.

% We find that \matrnn\ performs better than
% 	\sqrnn\ in the metrics considered,
% 	with \matrnn\ having a mean loss 
% 	[Equation \ref{eqn:phm08loss}]
% 	of 40.09
% 	compared to \sqrnn\ which has 193.36.
% In the RMSE metric (root-mean-squared-error),
% 	\matrnn\ had an error of 35.65
% 	compared to \sqrnn\ which as 36.48.
% It seemed that \matrnn\ is more biased towards under-estimating RUL
% 	which makes it perform much better in the custom loss metric.
% Also, we find that from the histogram of errors that
% 	\matrnn\ predictions is unimodal and clustered tightly around its mode
% 	[Figure \ref{fig:testing_cmapss_hist}].

% \begin{figure}[h]
% \centering
% \subfigure
% 	[Errors for \sqrnn]
% 	{\includegraphics[height=4cm]{pics/testing_cmapss_sqrnn_hist.png}}
% \subfigure
% 	[Errors for \mwtrnn]
% 	{\includegraphics[height=4cm]{pics/testing_cmapss_mwtrnn_hist.png}}
% \caption{Errors for \mwtrnn\ Mode and \sqrnn\ Estimate}
% \label{fig:testing_cmapss_hist}
% \end{figure}

% \begin{table}[h]
% \caption{Comparison on RNN based Methods for CMAPSS}
% \begin{center}
% \begin{tabular}{|l|r|r|r|r|}
% \hline
% \textbf{Model}
% 	& $W$
% 	& \textbf{MSE} & \textbf{RMSE} & \textbf{MCL} \\
% \hline
% \texttt{RNN}
% 	& 64
% 	& 1331.03 & 36.48 & 193.36 \\
% \matrnn
% 	& 64
% 	& {\bf 1271.49} & {\bf 35.65} & {\bf 40.09} \\
% \hline
% \end{tabular}
% \label{table:cmapssloss}
% \end{center}
% \end{table}

\subsection{Demand Forecasts for a Large Retailer}

We investigate the predictive performance 
	for purchases of of a few baskets of goods
	sold by a large retailer.
The time resolution of our dataset is on a weekly level.
Training data is available over roughly 1.5 years,
	which gives us 78 weeks of training data
	from 2014-01-01 to 2015-06-30.
Performance of our \mwtrnn\ model
	is measured on the binary classification problem
	of predicting whether a customer purchases the product 
	within 4 weeks after the end of training period
	from 2015-06-30 to 2015-07-31.

\subsubsection{Benchmark Models}

Even though our model
	can predict many different quantities of interest
	[Section \ref{predictions}],
	we chose to compare the predictive performance
	to a few benchmark models
	in terms of whether an event will arrive
	within $\gamma$ time after the end of the training period.
These models are namely the 
	Squared-Loss RNN (\texttt{SQ-RNN})
	and a Random Forest Predictor (\rngf),
	which are state-of-the-art models
	in personalized demand forecasting
	\cite{berry2004}
	and forecasting event arrival times
	\cite{choi2015}.
Models are trained on all customers who bought an item in the basket
	during the training period
	and performance is evaluated on this group of customers
	during the testing period.

\begin{itemize}

	\item \rngf\ is trained by 
		splitting the training period
		into two periods.
		Covariates at the end of the first period
			are fed into the model,
			which is trained to predict whether 
			subjects purchase in the second period
			of length $\gamma$.
		A different \rngf\ model is trained for each product,
			but is fed covariate datasets for all products.

	\item \sqrnn\ is trained by 
		setting the loss function 
		as the squared difference
		between the predicted time-to-arrival
		and the actual time-to-arrival.
		An activation function of softplus is applied.
		Predictions of \sqrnn\ are then 
			compared to the testing period length of $\gamma$.
		If by the end of the training period,
			\sqrnn\ predicts the next time-to-arrival as $s$,	
			then the prediction metric is $\gamma-s$.
		For time periods where no actual time-to-arrival was observed
			(i.e. no further purchases were observed by end of training),
			loss is set to 0.

	\end{itemize}

\subsubsection{Covariates}

For each customer, at each time period,
	we compute the Recency, Frequency and Monetary (RFM) metrics
	which are commonly used in demand modeling
	\cite{berry2004}
	at three different levels:
	namely for all products,
		in-basket products
		and each individual product.
Recency is the time since last purchase,
	frequency is the number of repeat purchases
	and monetary is the amount of money spent on all purchases to date.
Included in the covariates 
	are time-since-event ($\tse(t)$) and
	indicators for whether a first purchase has occurred ($\text{pch}(t)$).
We also compute the time-to-event ($\tse(t)$) as well as 
	the censoring status of the next arrival ($\text{unc}(t)$),
	which are only passed to the loss function.

On a per-product level, the types of covariates are limited
	to only RFM metrics (3 covariates)
	and transformations of purchase history (2 series).
RFM metrics on the category and overall purchase history levels
	are available as well,
	but these account for an additional 6 covariates
	that are shared across the various purchase arrival processes.
The total number of covariates for each product is thus 11,
	of which 6 are shared with other products.

\subsubsection{Data Summaries for Product Baskets}

We selected 5 baskets of popular replenishable products.
These are selected from products ranked by a score,
	where $N_\text{unique}$ is the number of unique customers
	and $X$ is the average purchases per customer.
\begin{equation}
\text{score} = X * \log N_\text{unique}
\label{eqn:basketselectionscore}
\end{equation}

The selected baskets are 
	\texttt{bars},
	\texttt{deli},
	\texttt{floss},
	\texttt{pads},
	\texttt{soda}.
Their data summaries are presented 
	[Table \ref{table:blag-datasummary}],
	where 
	$\mu_{\text{overall}}$ 
	is the average in-basket purchase counts,
	$\mu_{\text{per-sku}}$
	is the mean over the per-product average purchase counts 
	and
	$p_\text{others}$
	is the mean over the per-product proportion of buyers 
	who bought another product in-basket.
Also note that
	$p_\text{trial}$
	is the mean over the per-product proportion of trial customers
	(i.e. those who have made only a single purchase).

\begin{table}[h]
\caption{Data Summary of Product Baskets}
\begin{center}
\begin{tabular}{|c|rr|rr|rr|}
	\hline
		&	& \textbf{customers}
		&	&	
		&	& \\
	\textbf{basket}
		& \textbf{SKUs} & \textbf{(x1000)}
		& $\mu_\text{overall}$ & $\mu_\text{per-sku}$ 
		& $p_\text{others}$ & $p_\text{trial}$ \\
	\hline
	bars
		& 6 & 44
		& 4.78 & 0.79 
		& 0.71 & 0.43 \\
	deli
		& 12 & 79
		& 3.58 & 0.29 
		& 0.55 & 0.62 \\
	floss
		& 11 & 200
		& 2.58 & 0.23
		& 0.40 & 0.64 \\
	pads
		& 7 & 317
		& {\bf 2.26} & 0.32 
		& {\bf 0.28} & {\bf 0.66} \\
	soda
		& 8 & 341
		& 2.97 & 0.37 
		& 0.45 & 0.63\\
	\hline
	\end{tabular}
\label{table:blag-datasummary}
\end{center}
\end{table}

We note that \texttt{pads}
	has the highest proportion of trial customers
	along with the smallest proportion of customers
	who bought another item in the basket.
On the other hand,
	we find that $\mu_\text{per-sku}$ is roughly median 
	in the baskets considered.
This is similar for \texttt{floss} as well.
% It would be reasonable to expect
% 	that purchase patterns for one product
% 	should be strongly correlated 
% 	with purchase patterns for other products 
% 	in these categories
% 	since we can expect customers to stick to one product
% 	or not make any future purchases.
For these categories,
	it would be reasonable to expect
	product purchases are strongly dependent.
A good joint-prediction model 
	should separate trial purchasers from 
	repeat purchasers who decided to stick to one product 
	after trying another.

\subsubsection{Performance of Joint Predictions}

Performance is measured based on the ROC-AUC metric
	where each of the models
	(i.e. \rngf, \sqrnn, \mwtrnn)
	predict whether customers who made in-basket purchases
	will make another in-basket purchase in a 4 week period
	after the end of a training period of 78 weeks.
Framing this as a binary prediction problem
	allows us to compare these state-of-the-art models.
The RNN-based models share the same network specifications
	with $W=36$ [Section \ref{nnetspecs}]
	and predict arrival times jointly over different products
	for each customer.
These models are run for 100 iterations with a learning rate set to $1e-3$.
The \rngf\ model is trained with 100 trees
	with covariates at week 74
	and purchases between week 74 and 78
	but predicts purchases for only one product at a time
	using default parameters in $\texttt{scikit-learn}$.
As such, a separate \rngf\ model is trained for each product.
We note that \rngf\ model is favored
	since the training metric
	is directly related to the testing metric,
	whereas the RNN-based models
	train and predict based on arrival times.

%%%%%%%%%%%%%%%%%%%%%%%%%%%%%%%%%%%%%%%%%%%%%%%%%%%%%%%%%%%%%%%%%%%%%%%%%%%%%%%%
\begin{table*}[h]
\begin{center}
\caption{ROC-AUC for Products in Category.
	Empirical quantiles are taken over
		per-product ROC-AUCs in each category.}

\begin{tabular}{|c|r|r|r|r|rrrrr|r|}
\hline
	& \textbf{Customers}
	& 
	& 
	& \textbf{\# Improved}
	& \multicolumn{5}{|c|}{\textbf{ROC-AUC Quantiles}} 
	& \textbf{ROC-AUC} \\
\cline{6-10}
\textbf{Category} 
	& \textbf{(x1000)} % customers...
	& \textbf{Products}
	& \textbf{Model}
	& \textbf{over \rngf} % Count of Improvments
	& \textbf{Min} & \textbf{Q25} & \textbf{Q50} & \textbf{Q75} & \textbf{Max}
	& \textbf{Average} \\
\hline

bars
	& 44
	& 6
	& \rngf
	& -
	& {\bf 0.7696} & {\bf 0.7986} & {\bf 0.8428} & {\bf 0.8648} & {\bf 0.8710}
	& {\bf 0.8304} \\
&&
	& \sqrnn
	& 0
	& 0.6608 & 0.7165 & 0.7228 & 0.7406 & 0.7550
	& 0.7204 \\
&&
	& \matrnn
	& {\bf 2}
	& 0.7588 & 0.7762 & 0.8174 & 0.8537 & 0.8783
	& 0.8167 \\
\hline

% \textbf{Category} 
% 	& \textbf{(x1000)} % customers...
% 	& \textbf{Products}
% 	& \textbf{Model}
% 	& \textbf{over \rngf} % Count of Improvments
% 	& \textbf{Min} & \textbf{Q25} & \textbf{Q50} & \textbf{Q75} & \textbf{Max}
% 	& \textbf{Average} \\
deli
	& 79
	& 12
	& \rngf
	& -
	& 0.7452 & 0.7995 & 0.8389 & {\bf 0.9004} & {\bf 0.9220}
	& 0.8468 \\
&&
	& \sqrnn
	& 4
	& 0.7763 & 0.8047 & 0.8248 & 0.8458 & 0.8810
	& 0.8259 \\
&&
	& \matrnn
	& {\bf 8}
	& {\bf 0.8686} & {\bf 0.8823} & {\bf 0.8911} & 0.9021 & 0.9131
	& {\bf 0.8919} \\
\hline

% \textbf{Category} 
% 	& \textbf{(x1000)} % customers...
% 	& \textbf{Products}
% 	& \textbf{Model}
% 	& \textbf{over \rngf} % Count of Improvments
% 	& \textbf{Min} & \textbf{Q25} & \textbf{Q50} & \textbf{Q75} & \textbf{Max}
% 	& \textbf{Average} \\
floss
	& 200
	& 11
	& \rngf
	& -
	& 0.5537 & 0.6066 & 0.6199 & 0.6517 & 0.7683
	& 0.6408 \\
&&
	& \sqrnn
	& 10
	& 0.7298 & 0.7809 & 0.8089 & 0.8366 & 0.8739
	& 0.8055 \\
&&
	& \matrnn
	& {\bf 11}
	& {\bf 0.8680} & {\bf 0.9016} & {\bf 0.9317} & {\bf 0.9421} & {\bf 0.9640}
	& {\bf 0.9214} \\
\hline

% \textbf{Category} 
% 	& \textbf{(x1000)} % customers...
% 	& \textbf{Products}
% 	& \textbf{Model}
% 	& \textbf{over \rngf} % Count of Improvments
% 	& \textbf{Min} & \textbf{Q25} & \textbf{Q50} & \textbf{Q75} & \textbf{Max}
% 	& \textbf{Average} \\
pads
	& 317
 	& 7
	& \rngf
	& -
	& 0.5851 & 0.6148 & 0.6358 & 0.6411 & 0.8234
	& 0.6509 \\
&&
	& \sqrnn
	& 4
	& 0.5650 & 0.6149 & 0.6392 & 0.6941 & 0.7154
	& 0.6482 \\
&&
	& \matrnn
	& {\bf 7}
	& {\bf 0.8544} & {\bf 0.9160} & {\bf 0.9459} & {\bf 0.9511} & {\bf 0.9621}
	& {\bf 0.9281} \\
\hline

% \textbf{Category} 
% 	& \textbf{(x1000)} % customers...
% 	& \textbf{Products}
% 	& \textbf{Model}
% 	& \textbf{over \rngf} % Count of Improvments
% 	& \textbf{Min} & \textbf{Q25} & \textbf{Q50} & \textbf{Q75} & \textbf{Max}
% 	& \textbf{Average} \\
soda
	& 341
 	& 8
	& \rngf
	& -
	& 0.6959 & 0.7372 & 0.7663 & 0.7903 & 0.8300
	& 0.7641 \\
&&
	& \sqrnn
	& 1
	& 0.6844 & 0.7221 & 0.7259 & 0.7320 & 0.7612
	& 0.7258 \\
&&
	& \matrnn
	& {\bf 8}
	& {\bf 0.8605} & {\bf 0.8669} & {\bf 0.8795} & {\bf 0.8854} & {\bf 0.8909}
	& {\bf 0.8768} \\
\hline

\end{tabular}

\label{table:blag-rocauc}
\end{center}
\end{table*}
%%%%%%%%%%%%%%%%%%%%%%%%%%%%%%%%%%%%%%%%%%%%%%%%%%%%%%%%%%%%%%%%%%%%%%%%%%%%%%%%

We are interested in the performance of each model 
	for every product in the basket
	so there are multiple ROC-AUC metrics.
The results are presented in terms of summary statistics
	for ROC-AUCs for each item in the basket
	[Table \ref{table:blag-rocauc}].
In our testing, we found that the \mwtrnn\ model
	almost always dominates in the ROC-AUC metric
	for every category other than \texttt{bars} and \texttt{deli},
	which has the smallest number of customers.
Even so, \mwtrnn\ still performs the best
	in terms of average ROC-AUC among products in each category
	[Table \ref{table:blag-rocauc}]
	other than \texttt{bars}.

The number of products 
	for which ROC-AUC has improved over \rngf\
	is substantial for \mwtrnn.
Excluding \texttt{bars}
	where only 2 out of 6 products saw improved performance,
	other categories saw ROC-AUC improvements 
	in more than 60\% of the products in-category,
	with \texttt{soda} and \texttt{pads} showing improvements in all products.
The ability to model sequential data 
	and sequential dependence
	separates \matrnn\ model from \rngf.
Even though \rngf\ is trained on the evaluation metric,
	we find that \matrnn\ almost always performs better
	in this binary classification task.

% especially for pads... address why
Also notable is that the 
	performance difference of \mwtrnn\ over \sqrnn\ and \rngf\
	is greatest for the \texttt{pads} category.
This is likely due to the large amount of missing data
	since customers are least likely to buy other products
	[Table \ref{table:blag-datasummary}].
% sqloss doing worse than mwtrnn
We also find that \sqrnn\ performs poorly compared to \matrnn\
	[Table \ref{table:blag-rocauc}],
	even though these models have the same recurrent structure
	and are fed the same sequential data.
One possible explanation is that the lack of ground truth data
	has a significant impact on the ability of \sqrnn\ to learn.
In cases where event arrivals are sparse
	or where inter-purchase periods are long,
	the censored nature of the data
	gives no ground truth to train \sqrnn\ on.
Therefore, even though the recurrent structure
	makes it possible to model sequential dependence,
	the structure that \matrnn\ imposes on the problem
	makes it much easier to make predictions with censored observations.

% effect of sample size
We should note that \rngf\ out-performs \mwtrnn\
	for 4 out of 12 \texttt{deli} products
	and 4 out of 6 \texttt{bars} products.
A possible reason is that these categories have a smaller sample size.
One limitation of the RNN models
	is that regularization options are more limited
	compared to ensemble methods used in \rngf.
Based on the results, 
	it appears that \mwtrnn\ performs better
	for more customers
	[Table \ref{table:blag-rocauc}].

% % floss 0.24
% Also \texttt{floss} has one product that the RNN-based models
% 	performed poorly on.
% Examining a plot of weekly purchase counts of
% 	\texttt{Sku0} against other products reveal that
% 	repeat purchases did not occur 
% 	during the first third of the training period
% 	[Figure \ref{fig:euroretailer_flossdiagnosis}].
% This contributed to \texttt{floss} having the smallest
% 	$\mu_\text{per-sku}$ 
% 	[Table \ref{table:blag-datasummary}].
% For this sku, it appears that the lack of consistent repeat purchases
% 	biased the RNN-based models towards
% 	over-fitting and predict low purchase probabilities.

% \begin{figure}[h!]
% 	\centering
% 	\includegraphics[height=5cm]{pics/euroretailer_flossdiagnosis.png}
% 	\caption{Weekly Repeat Purchase Counts of \texttt{Floss} Products}
% 	\label{fig:euroretailer_flossdiagnosis}
% \end{figure}

\subsubsection{Joint Predictions and Individual Predictions}

It is clear that joint predictions enjoy some advantages
	over individual predictions.
As stated earlier, we expect correlations in product purchases
	to be modeled better through joint modeling.
If network structure is the same,
	then the amount of time required to train a separate model 
	for each product scales linearly with the number of products.
The number of parameters in a collection of individual models
	is also significantly larger than that of a joint model.

We study the advantages of training a joint \mwtrnn\ model
	over a collection of individual ones
	by comparing ROC-AUC performance 
	in the \texttt{soda} basket.
The per-product individual models
	are given the same covariates
	but trained only on the purchase arrivals
	of that particular product.
The network structure is the same with $W=36$,
	but the final densely connected layer
	outputs only a vector of size 2,
	since distributional parameters for one product is required.
However, since the collection of single models
	have different weights for their RNNs,
	they have approximately 8 times the number of parameters
	found in the joint model.
We observe a consistent advantage
	of a joint model over the individually trained single models,
	with improvements ranging from 0.0029 to 0.1098
	[Table \ref{table:blag-soda-indivjoint}].
It is clear that potential improvements in model performance
	can be observed by modeling purchase arrivals jointly,
	even with much fewer parameters in the joint model.

\begin{table}[h!]
\caption{Comparison of ROC-AUC performance 
	on \texttt{soda} 
	for single and joint \mwtrnn\ models}
\begin{center}
\begin{tabular}{|c|rrr|}
	\hline
	\textbf{sku} & \textbf{single} & \textbf{joint} & \textbf{diff} \\
	\hline
	1 & 0.8868 & {\bf 0.8897} & +0.0029 \\
	2 & 0.8073 & {\bf 0.8686} & +0.0614 \\
	3 & 0.8331 & {\bf 0.8605} & +0.0274 \\
	4 & 0.8501 & {\bf 0.8761} & +0.0260 \\
	5 & 0.8445 & {\bf 0.8829} & +0.0384 \\
	6 & 0.8193 & {\bf 0.8615} & +0.0422 \\
	7 & 0.8640 & {\bf 0.8909} & +0.0269 \\
	8 & 0.7742 & {\bf 0.8840} & +0.1098 \\
	\hline
	\end{tabular}
\label{table:blag-soda-indivjoint}
\end{center}
\end{table}

\section{Conclusion}
We described a method
	to incorporate a survival analysis approach
	with recurrent neural nets (RNN)
	to forecast joint arrival times till next purchase
	for each individual customer
	over multiple products.
This was achieved by transforming the arrival time problem
	into a likelihood-maximization one
	with loose distributional assumptions
	regarding inter-arrival times.
Our experiments show that
	this leads to significant improvement over
	current state-of-the-art methods is possible.
This is the result of being able to model purchases jointly
	as well as combining a parametric approach 
	to modeling partially observed information
	with recent advances in neural networks.

\section{Future Work}
Future work includes modeling more features,
	as the structure allows for the observed processes
	to be independent given the parameter generation process.
An extra feature to model is first-arrival times,
	which can help predict whether a customer might make their first purchase
	in response to marketing campaigns.
However, to model this effectively,
	we require a model-based solution 
	to address the sparsity of observations of first-arrival times.

\subsection{Scaling to Inventory-Wide Predictions}

Including thousands of products is infeasible in the current implementation.
It would be useful to extend the model to make 
	joint inventory-wide predictions for purchase arrival times.
Inventory-wide predictions also brings about other technical problems such as 
	memory usage and ETL (data extraction, transformation, loading).
The computation of time-since-event and time-to-event
	as well as other features to be used in log-likelihood computation
	[Section \ref{llikecomp}]
	is also problematic in that it transforms
	a sparse dataset (i.e. transactional information) into
	a dense dataset (i.e. temporal data).
An inventory-wide extension needs to address these issues
	by either implementing an efficient method to compute these matrices
	or avoid the problem through a variant of \matrnn.
Nonetheless, we believe that the current implementation
	can suffice for the purposes of targeted advertising 
	where we are only interested in 
	predicting demand and pushing sales for a few product lines.
%%%%%%%%%%%%%%%%%%%%%%%%%%%%%%%%%%%%%%%%%%%%%%%%%%%%%%%%%%%%%%%%%%%%%%%%%%%%%%%%

%%%%%%%%%%%%%%%%%%%%%%%%%%%%%%%%%%%%%%%%%%%%%%%%%%%%%%%%%%%%%%%%%%%%%%%%%%%%%%%%
\section{Acknowledgments}

We would like to thank Kanchana Padmanabhan
	and Erin Wilder
	at Rubikloud Technologies Inc
	for their expertise in editing this paper
	as well as the Data Science and Engineering team
	for their technical support.
This would also not have been possible
	without the supervision and advice 
	of Professor Nancy Reid and Professor Andrei Badescu
	at the University of Toronto.
Lastly, we'd like to thank 
	the NSERC (Natural Sciences and Engineering Research Council) Engage and 
	Mitacs Accelerate programs
	for funding this project.
%%%%%%%%%%%%%%%%%%%%%%%%%%%%%%%%%%%%%%%%%%%%%%%%%%%%%%%%%%%%%%%%%%%%%%%%%%%%%%%%

%\appendix

\section{\label{appendix:weibull_survival} Appendix: Weibull Likelihoods}

A random variable $Y\sim\texttt{Weibull}(\textbf{scale}=\lambda,\textbf{shape}=k)$ 
	has simple densities and cumulative distribution functions, 
	since the survival function $(S(x))$ has a simple form:

\begin{equation}
\begin{aligned}
	S(y) 
		&= P(Y>y) \\
		&= e^{-(y/\lambda)^k}. \\ 
	f(y) 
		&= (k/\lambda)(y/\lambda)^{k-1} e^{-(y/\lambda)^k} \\
		&= (k/\lambda)(y/\lambda)^{k-1} S(y). \\
	\end{aligned}
\end{equation}

The conditional excess random variable, 
	given that it exceeds $s$, is $W=Y-s|Y>s$.
Recall the definition of conditional probability
	in terms of some continuous random variable $X_1, X_2$,
	for any measurable set $A_1, A_2$,
	given $P(X_2 \in A_2) > 0$:

\begin{equation}
P(X_1 \in A_1 | X_2 \in A_2)
	= \frac{P(X_1 \in A_1, X_2 \in A_2)}{P(X_2 \in A_2)}.
\end{equation}

We can therefore derive the conditional excess survival function:

\begin{equation}
\begin{aligned}
	S_W(t)
		&= P(W>t) \\
		&= P(Y>s+t|Y>s) \\
		&= S(s+t) / S(s) \\
		&= \exp \left\{
			-((s+t)/\lambda)^k 
			+(s/\lambda)^k 
			\right\}. \\ 
	\end{aligned}
\end{equation}

Also, we can find the conditional excess density function:

\begin{equation}
\begin{aligned}
	f_W(t)
		& = f(s+t) / S(s) \\
		& = (k/\lambda)((s+t)/\lambda)^{k-1} S(s+t) / S(s) \\
		& = (k/\lambda)((s+t)/\lambda)^{k-1} S_W(t). \\
	\end{aligned}
\end{equation}

\bibliography{_bibliography}
\bibliographystyle{plain}

\end{document}